\pdfoutput=1

\documentclass[11pt]{article}

\usepackage[review]{acl}

\usepackage{times}
\usepackage{latexsym}

\usepackage[T1]{fontenc}

\usepackage[utf8]{inputenc}

\usepackage{microtype}

\usepackage{inconsolata}

\usepackage{graphicx}
\usepackage{booktabs}
\usepackage{amsmath}
\usepackage{placeins}
\usepackage{xcolor}
\usepackage{multirow}
\usepackage{enumitem}
\usepackage{hyperref}
\usepackage{makecell}
\DeclareSymbolFont{extraup}{U}{zavm}{m}{n}
\DeclareMathSymbol{\varheart}{\mathalpha}{extraup}{86}
\DeclareMathSymbol{\vardiamond}{\mathalpha}{extraup}{87}
\usepackage{pifont}
\newcommand{\xmark}{\ding{55}\,\,}
\usepackage{amsmath}

\title{Language Models as Inductive Reasoners}

\author{
    Zonglin Yang\textsuperscript{\rm $\spadesuit$}\thanks{\,\,\,Contribution during internship at Microsoft Research.}\,\,
    Li Dong\textsuperscript{\rm $\diamondsuit$}
    Xinya Du\textsuperscript{\rm $\clubsuit$}
    Hao Cheng\textsuperscript{\rm $\diamondsuit$} \\
    \textbf{Erik Cambria}\textsuperscript{\rm $\spadesuit$}
    \textbf{Xiaodong Liu}\textsuperscript{\rm $\diamondsuit$}
    \textbf{Jianfeng Gao}\textsuperscript{\rm $\diamondsuit$}
    \textbf{Furu Wei}\textsuperscript{\rm $\diamondsuit$} \\
    \textsuperscript{\rm $\spadesuit$} {\normalsize Nanyang Technological University}
    \textsuperscript{\rm $\diamondsuit$} {\normalsize Microsoft Research} \\
    \textsuperscript{\rm $\clubsuit$} {\normalsize University of Texas at Dallas} \\
    {\tt \normalsize \{zonglin.yang,cambria\}@ntu.edu.sg} \\
    {\tt \normalsize \{lidong1,cheng.hao,xiaodl,jfgao,fuwei\}@microsoft.com} \\
    {\tt \normalsize xinya.du@utdallas.edu}
}

\begin{document}
\maketitle
\begin{abstract}
Inductive reasoning is a core component of human intelligence.
In the past research of inductive reasoning within computer science, formal language is used as representations of knowledge (facts and rules, more specifically).
However, formal language can cause systematic problems for inductive reasoning such as disability of handling raw input such as natural language, sensitiveness to mislabeled data, and incapacity to handle ambiguous input.
To this end, we propose a new paradigm~(task) for inductive reasoning, which is to induce natural language rules from natural language facts, and create a dataset termed DEER containing 1.2k rule-fact pairs for the task, where rules and facts are written in natural language.
New automatic metrics are also proposed and analysed for the evaluation of this task.
%
With DEER, we investigate a modern approach for inductive reasoning where we use natural language as representation for knowledge instead of formal language and use pretrained language models as ``reasoners''.
%
Moreover, we provide the first and comprehensive analysis of how well pretrained language models can induce natural language rules from natural language facts.
We also propose a new framework drawing insights from philosophy literature for this task, which we show in the experiment section that surpasses baselines in both automatic and human evaluations.
We discuss our future perspectives on inductive reasoning in detail in Section~\ref{sec:overview_future_perspectives}.
Dataset and code are available at \url{https://github.com/ZonglinY/Inductive_Reasoning}.
\end{abstract}

\section{Introduction}

Inductive reasoning is to reach to a hypothesis (usually a rule that explains an aspect of the law of nature) based on pieces of evidence (usually observed facts of the world), where the observations can not provide conclusive support to the hypothesis~\citep{salmon1989introduction}. 
It is ampliative, which means that the hypothesis supports more than mere reformulation of the content of the evidence~\citep{norton2005little}.   
An example is shown in Table~\ref{tab:examples_inductive_reasoning} that after observing three carnivorous plants each having a trapping structure, one might reach to a hypothesis~(rule) that every carnivorous plant has a trapping structure.
%
Inductive reasoning was firstly proposed by Aristotle in the 4th century B.C. in his {\it Posterior Analytics}~\citep{aristotle1994posterior}. 
Since then it is used as a fundamental tool to obtain axioms, and therefore subjects can be developed from these axioms. 
It is also recognized as a core component of human intelligence~\citep{mercier2018enigma}.



\begin{table*}[]
\centering
\resizebox{2.0\columnwidth}{!}{
\begin{tabular}{cccc}
\toprule
Short fact 1  & Short fact 2  & Short fact 3    & Rule                                                                  \\ \midrule 
\thead{The Venus flytrap is a \textbf{carnivorous} \\\textbf{plant} native to subtropical wetlands\\ on the East Coast of the United States\\ in North Carolina and South Carolina.\\ It catches its prey-chiefly insects \\and arachnids—with a \textbf{trapping structure} \\formed by the terminal portion of each \\of the plant's leaves,  which is triggered \\by tiny hairs on their inner surfaces.}
& \thead{Pitcher plants are several different\\ \textbf{carnivorous plants} which have modified\\ leaves known as \textbf{pitfall traps}—a prey\\-trapping mechanism featuring a deep \\cavity filled with digestive liquid. The \\traps of what are considered to be "true" \\pitcher plants are formed by \\specialized leaves. The plants attract \\and drown their prey with nectar.} 
& \thead{Drosera, which is commonly known \\as the sundews, is one of the largest genera \\of \textbf{carnivorous plants}, with at least\\ 194 species. The \textbf{trapping} and digestion \\mechanism of Drosera usually employs \\two types of glands: stalked glands that \\secrete sweet mucilage to attract and ensnare\\ insects and enzymes to digest them, and sessile\\ glands that absorb the resulting nutrient soup.} 
& \thead{If a \\plant is\\ carnivorous\\, then it\\ probably \\has a\\ trapping\\ structure.} \\ \bottomrule
\end{tabular}
}

\caption{An example of inductive reasoning in DEER dataset.
We embolden the words in facts that contain the key information to induce this rule~(just to explain the relation between facts and rule, in DEER there's no special word annotations for fact).
}

\label{tab:examples_inductive_reasoning}
\end{table*}

Past research works on inductive reasoning within computer science are investigated by Inductive Logic Programming~(ILP)~\citep{DBLP:journals/ml/MuggletonRPBFIS12}.
ILP investigates the inductive construction of first-order logic~(FOL)~\citep{smullyan1995first} rules from examples and background knowledge~\citep{DBLP:journals/jlp/MuggletonR94}. 
%
However, ILP uses formal language as representation and uses symbolic 
reasoner, which results in systematic disadvantages~\citep{DBLP:journals/ml/CropperDEM22}. 
Specifically, ILP systems heavily rely on human effort, since it typically assumes that the input has already been preprocessed into symbolic declarative form, otherwise ILP systems cannot handle raw inputs such as natural language and images. 
In addition, ILP systems are very sensitive to label error and ambiguity in data, since the final induced rules are required to satisfy all input facts, and symbolic systems can not recognize different symbols with the same meaning (e.g. be capable of, be able to).

To overcome the challenges above, we present a novel paradigm for inductive reasoning based entirely on natural language, i.e., inducing natural language rules from natural language facts. 
In particular, we create a first-of-its-kind natural language inductive reasoning dataset named DEER containing 1.2k rule-fact pairs~(more details illustrated in \S\ref{subsection:data_collection_DEER}).
%
With this dataset, we investigate a modern approach to inductive reasoning
where both facts and rules are in natural language, and pretrained language models~(PLMs) are used as the inductive reasoner. 
Note that the inductive reasoning considered in this paper has several distinctions considered by other reasoning tasks over text~\citep{DBLP:conf/ijcai/ClarkTR20,DBLP:conf/iclr/BhagavatulaBMSH20,DBLP:conf/emnlp/SinhaSDPH19}. We defer a more detailed discussion to \S\ref{ssec:relation_to_others}.

With natural language as representation and PLMs as the reasoner, 
such an 
inductive reasoning system can avoid the systematic disadvantages of formal language and symbolic reasoners.
Specifically, with natural language as representation, it can naturally handle raw input as natural language text.
In addition, different from symbolic methods, PLMs contain knowledge via pretraining~\citep{DBLP:conf/emnlp/DavisonFR19} and use embedding for concepts~\citep{DBLP:conf/nips/MikolovSCCD13}, making it less affected by input errors~\citep{DBLP:conf/emnlp/0001ZHWZJ021} and more robust to paraphrasing.




Based on the proposed dataset,
we study the PLM's ability to induce (generate) natural language rules from natural language facts under different settings, such as different FOL rule types and topics
with varying input facts and PLM model sizes.


We also propose a new framework for this task, named chain-of-language-models~(CoLM) which is shown in Figure~\ref{fig:main_framework}.
It draws insights from the requirements of rule induction in philosophy literature~\citep{norton2005little}.
Specifically, CoLM consists of five modules all based on PLMs, where one model proposes rules (rule proposer M1), and the other four models (M2, M3, M4, M5) each classify whether a generated rule satisfies one particular requirement of induction.  
In our experiments, we find that our framework surpasses the baselines in terms of both automatic and human evaluations.

To sum up, our contributions are three-fold:
\begin{itemize}
    \item We propose a new paradigm~(task) of inducing natural language rules from natural language facts, which naturally overcomes three systematic disadvantages of past works on inductive reasoning. 
    In particular, we create a first-of-its-kind natural language inductive reasoning dataset DEER containing 1.2k rule-fact pairs, where fact and rule are both written in natural language.
    New automatic metrics are also proposed for task evaluation, which shows strong consistency with human evaluation.
    \item We provide the first and comprehensive analysis of how well PLMs can induce natural language rules from natural language facts.
    \item Drawing insights from philosophy literature~\citep{norton2005little}, we propose a framework for inductive reasoning.
    Empirically, we show that it surpasses baselines substantially in both automatic and human evaluations.

\end{itemize}

In \S\ref{sec:overview_future_perspectives} we discuss our future perspectives on inductive reasoning in detail.

\section{Related Work}


\paragraph{Definition of Inductive Reasoning}
It is still under debate on the definition of inductive reasoning in philosophy research~\citep{DBLP:journals/corr/abs-2303-12023}.
Here we adopt~\citet{flach2000abductive}'s view that an inductive argument should satisfy
(1) its premise cannot provide conclusive support to its conclusion since its conclusion amplify or go beyond the information found in their premises;
(2) its conclusion generalize over its premise in a way that the conclusion can be applied to more instances other than instances mentioned in its premise.
An example of inductive argument is that ``if a white ball is found in a bag, then all balls in this bag are white.''
%
%
In this paper, we call the premises as ``facts'', and conclusions as ``rules''.
Prior computational method for inductive reasoning is inductive logic programming, which is introduced in \S\ref{appen:inductive_logic_programming}.

\paragraph{Inductive Reasoning \& Neural Networks}
\citet{DBLP:conf/emnlp/SinhaSDPH19} propose CLUTRR dataset, 
but they do not focus on inducing explicit natural language rules.
Instead they try to ``learn'' certain rules internally with PLMs, and use the PLMs to predict the correctness of other facts.
Inductive relation induction task~\citep{DBLP:conf/icml/TeruDH20,DBLP:journals/corr/abs-2205-06910} focuses on prediction of relation that involves unseen entities, which only involves an induction from specific entities to specific entities, where we focus on the induction from specific entities or individual phenomenons to general knowledge.
\citet{DBLP:journals/corr/abs-2111-12038} also works on rule induction, but their induced rule is not in real natural language, and uses symbolic reasoners.



\paragraph{Relation with Other Reasoning Tasks}
\label{ssec:relation_to_others}
The goal is quite different from (1) deductive reasoning as given facts and rules and reach to new facts~\citep{DBLP:conf/ijcai/ClarkTR20}
(2) abductive reasoning as given facts and finding the casual reasons~\citep{DBLP:conf/iclr/BhagavatulaBMSH20}.
Rather, we want to induce rules that generalize over facts.
\citet{DBLP:journals/corr/abs-2303-12023} provide a comprehensive discussion on the difference between deductive, inductive, and abductive reasoning.
%
 
\begin{table}[]
\resizebox{\columnwidth}{!}{
\begin{tabular}{c|c}
\toprule
\thead{Rule Template\\(First Order Logic)} & \thead{Rule Template\\(Natural Language)} \\ \midrule
$\forall x$, $condition(x) \implies conclusion$  & If \_\_, then \_\_.            \\
$\exists x$, $condition(x) \implies conclusion$ & There exists \_\_, which \_\_.\\
\thead{$\forall x,$ $condition(x)$ [$\land$ $condition(x)$]$^{+}$\\ $\implies$ $conclusion$} & If \_\_ and \_\_, then \_\_.   \\
\thead{$\forall x,$ $condition(x)$ [$\lor$ $condition(x)$]$^{+}$\\ $\implies$ $conclusion$} & If \_\_ or \_\_, then \_\_.    \\
                    \bottomrule
\end{tabular}}
\caption{The mapping relation between basic first-order logic rule template and natural language rule template.}
\label{tab:rule_template}
\end{table}
\section{Dataset Collection and New Metrics}
\label{sec:dataset_collection_evaluation_metrics}

In this section, we discuss the data collection process for our proposed dataset, and our proposed metrics for automatic and human evaluation.

In general, we propose two datasets. 
The first one, named DEER~(in\textbf{D}uctive r\textbf{E}asoning with natural languag\textbf{E} \textbf{R}epresentation), contains 1.2k rule-fact pairs, where rules are written by human annotators in English, and facts are existing English sentences on the web.
%
The other one, named DEERLET (classification of in\textbf{D}uc\textbf{E}d rul\textbf{E}s with natu\textbf{R}al \textbf{L}anguag\textbf{E} representa\textbf{T}ion), including (fact, rule, label0, label1, label2, label3) tuples, where facts are the same as in DEER, rules are generated output from PLMs,
and label0/1/2/3 are classification labels describing different aspects of induced rules. 
Specifically, rules in DEERLET are collected from GPT-J~\citep{gpt-j} using the in-context learning setting.
We choose this setting because (1) GPT-J in this setting can generate reasonable rules, and (2) not all generated rules are correct so that the annotations on the generated rules can be used for fine-tuning.
%
Overall, DEER is used as the main dataset for the task, and DEERLET is used to measure the classification performance of specific capabilities described in \S\ref{subsec:deerlet_subsection}.

\subsection{Dataset Collection of DEER}
\label{subsection:data_collection_DEER}
Collected by a human expert~(the first author), DEER contains 1.2k natural language rule-fact pairs where rules cover 6 topics and 4 common rule types of FOL.
The 6 topics are zoology, botany, geology, astronomy, history, and physics. 
Shown in Table~\ref{tab:rule_template}, sequentially the 4 FOL rule types are implications with universal quantifier,
implications with existential quantifier,
conjunctive implications with universal quantifier,
and disjunctive implications with universal quantifier.
In practice we collect rules with the natural language rule templates.

Natural language rule is firstly written by a human expert,
then for each rule 6 supporting facts~(3 long facts and 3 short facts)
are collected from existing human-written text from commercial search engines and Wikipedia.
Long facts are paragraphs collected from different web pages to for more difference, and short facts are core sentences selected from corresponding long facts.
Each fact itself should contain enough information that is possible to induce the full corresponding rule~(an example is shown in Table~\ref{tab:examples_inductive_reasoning}).


To validate the correctness of the DEER dataset, we randomly split DEER data to 4 subsets, and 4 graduate students manually check each of the subsets on whether each fact contains enough information that is possible to induce the given rule.
The overall correctness of DEER is 95.5\%.

The reason that DEER is not larger is that it requires experts who are familiar enough with inductive reasoning and possesses a relatively high level of science knowledge to annotate. 

\subsection{Dataset Collection of DEERLET}
\label{subsec:deerlet_subsection}
DEERLET is a dataset collected by a human expert~(the first author) in inductive reasoning for classification tasks to evaluate the specific capabilities required by inductive reasoning. 
It contains 846 tuples with format (fact, rule, label0, label1, label2, label3). 
Among the tuples, 546 are used for training, 100 for validation, and 200 for testing.
Here, facts are directly from DEER, and the corresponding rules are collected from PLMs.
Label0 to label3 are classification labels evaluating specific aspects of the generated rules. 
The reason in DEERLET we collect rules from the generation of PLMs is that we want to avoid human annotation biases~\citep{DBLP:conf/coling/AmideiPW20}. 

We develop label 0/1/2 based on the requirements of induced rules in philosophy literature~\citep{norton2005little}, and develop label 3 based on a NLP aspect.
In particular, 
label0 measures whether a rule is not in conflict with its fact; 
label1 measures whether a rule reflects reality;
label2 measures whether a rule is more general than its fact, as inductive reasoning is ``ampliative'', and requires the induced rule to have higher coverage than facts~\citep{norton2005little}.
More details on label2 is illustrated in \S\ref{appen:general_inductive_reasoning}.
Label3 measures whether a rule is not trivial~(mostly incomplete sentence or the latter part is a repetition of its former part).

Inspired by~\citet{DBLP:conf/inlg/ObeidH20}, label 0/1/2 are annotated on a 3-point scale~(true / partially true / false), and label 3 are annotated on a 2-point scale~(true / false).
More details on annotation of DEERLET are illustrated in \S\ref{appen:annotation_detail_deerlet}.


\subsection{Adopted \& New Evaluation Metrics}
\label{adopted_proposed_automatic_evaluation_metrics}
\begin{table}[]
\resizebox{\columnwidth}{!}{
\begin{tabular}{c|cccc}
\toprule
      & \thead{Generated rules\\ with top \\0\%$\sim$top10\% \\METEOR} 
      & \thead{Generated rules\\ with top\\ 10\%$\sim$top20\% \\METEOR} & 
      ... 
      & \thead{Generated rules\\ with top\\ 90\%$\sim$top100\% \\METEOR} \\ \midrule
Weight & $weight_{0} (45)$                              & $weight_{1} (35)$  
        & ... 
        & $weight_{9} (-45)$                                               \\
Recall & $recall_{0}$                                        & $recall1_{1}$                                         & ... & $recall_{9}$     \\
\bottomrule
\end{tabular}}
\caption{Illustration of the weights and recalls in WRecall, one of our proposed automatic evaluation metrics. Here weights reflect the importance of blocks of rules.}
\label{tab:wrecall}

\end{table}

\begin{figure*}[t]
\centering
\resizebox{2\columnwidth}{!}{
\includegraphics[]{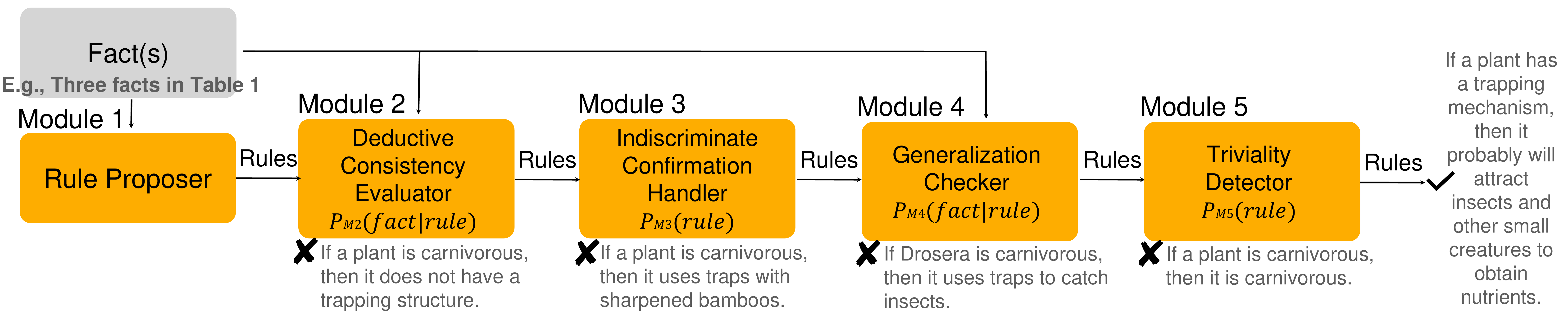}
}
\caption{Our proposed framework~(CoLM) for inductive reasoning with natural language representation task. 
Rule Proposer is a generative model based on input facts and desired rule template, aiming at generating (a large number of) rule candidates. 
Deductive consistency evaluator, indiscriminate confirmation handler, generalization checker, and triviality detector are classification models that filter improper rules according to four requirements of the induced rules in inductive reasoning.
Texts with \xmark are representative filtered rules for each module.
}
\label{fig:main_framework}
\end{figure*}

\subsubsection{Human Evaluation Metric}
DEERLET provides human annotations for evaluation of the generated rules from four different aspects.
Here we use precision / recall / f1, and the four aspects in DEERLET for human evaluation.

\subsubsection{Automatic Evaluation Metric}
For the DEER dataset, as it requires generating rules based on input facts, the first metric we adopt is METEOR~\citep{DBLP:conf/acl/BanerjeeL05}, which has been widely used for evaluating machine-generated text quality.
\S\ref{appen:why_meteor_not_bleu} compares METEOR and BLEU~\citep{DBLP:conf/acl/PapineniRWZ02}, and illustrates the reasons why METEOR should be a better metric for this task.
%
More specifically, we calculate the averaged METEOR score of the generated rules (after filtering, if a model had a filtering phase).
From the observation that even humans still constantly make mistakes on inductive reasoning, we assume any framework for this task might~(but not necessarily) contain two phases as generation and filtering to obtain higher performance.
However, if with a filtering phase, METEOR only considers the rules that are not filtered. 

It makes the METEOR metric here a similar metric to ``precision'', as it only calculates the score for rules that are classified as ``true''.
As a result, the model might have a low recall in that it might only keep the rule with the highest confidence score, and classify many reasonable good rules as ``false''.

To measure the ``recall'' of inductive reasoning models, we propose ``weighted recall~(WRecall)'' as the second automatic evaluation metric for this task. 
The difficulty lies in that we don't have the ground truth labels for generated rules without human evaluation.
To calculate WRecall, we make an assumption, which is that the higher METEOR a rule has, generally the higher probability it is a reasonable rule for given facts.
This assumption is reasonable given the relatively high correlation coefficient between METEOR and human evaluation shown in \S\ref{appen:why_meteor_not_bleu}.
Specifically, as shown in table~\ref{tab:wrecall},
we can first calculate the METEOR for each generated rule, and sort them based on the value of METEOR. 
Then we calculate the recall value for each block of generated rules, during which we assume only the rules in that block have ``true'' ground truth label.
We also add a linearly changing weight for each block according to their importance.
To ensure WRecall is in the range [0,1], WRecall is linearly normalized:

\begin{equation}
    \scriptsize
    W\!Recall = \frac{\sum_{i=0}^{9} weight_{i}*recall_{i} + 125}{250}
\end{equation}

Now that we have a METEOR metric that provides a similar measurement of ``precision'', and WRecall for ``recall'', we propose GREEN~(\textbf{G}eomet\textbf{R}ic m\textbf{E}an of M\textbf{E}TEOR a\textbf{N}d WRecall) to consider METEOR and WRecall together.
It is defined as a geometric mean instead of a harmonic mean because METEOR is not in the range [0, 1].
More specifically, 

\begin{equation}
    \scriptsize 
    GREEN = \sqrt{METEOR*W\!Recall}
\end{equation}

In general, compared with METEOR, GREEN gives a more comprehensive evaluation of the induced rules. 
Therefore GREEN can be a more favorable metric when the recall is an important factor (e.g., computational power is limited).
However, when the precision of the induced rules is more favored, METEOR should be a more proper metric than GREEN.
\S\ref{appen:meteor_or_green} discusses more on the importance of each metric for this task. 
More discussions on the usage of automatic evaluation metrics and how should we interpret the results of automatic metrics
can be found in \S\ref{appen:difficulty_automatic_metrics}.

\section{Methodology}

In this section, we formally present the task definition and our proposed framework for natural language inductive reasoning. 
Figure~\ref{fig:main_framework} illustrates the general architecture of our proposed approach.

\subsection{Task Definition}
DEER dataset is used as the dataset for the natural language inductive reasoning task. 
The data format for DEER is ($rule$, $fact$), where both $rule$ and $fact$ are natural language sentences.
The goal of the task is to generate reasonable natural language $rules$ given $fact$ in an inductive reasoning way~(the rules should be more general and therefore cover more information than $fact$).

\subsection{Our Framework}
\label{subsec:our_framework}

Hypothetical Induction is an important induction type in inductive reasoning~\citep{norton2005little}. 
It can be understood as when people make observations, they might propose a hypothesis as a general rule that can entail the observations.
For example, when people observe that the Sun rises and falls every day, they might induce a hypothesis that the Earth is rotating itself, which is more general than the observations as the hypothesis can also help to explain the observable movements of the other Milky Way stars relative to the Earth.

Hypothetical induction fits our task well, as in DEER we also want to induce a hypothesis as a more general rule that can entail the facts.
We borrow insights from the requirements for the induced rules in hypothetical induction to develop our framework.
Specifically, there are mainly three requirements~\citep{salmon1989introduction,norton2005little}.
The first is that a correct hypothesis should be able to entail deductively as many observations as possible. 
The second is that the hypothesis should follow the laws of nature, as one could always concoct some imaginary hypothesis that is able to explain the observations but violates reality~(e.g., the Earth is the center of the Universe so that the Sun orbits around the Earth).
In inductive reasoning, the failure to recognize a rule that runs counter to reality is called ``indiscriminate confirmation''.
The third is a basic requirement for inductive reasoning, where the hypothesis should be a more general statement than the observations~(Appendix~\ref{appen:general_inductive_reasoning} illustrates the meaning of ``general'').
We additionally introduce a fourth requirement from NLP aspects since this task uses natural language as knowledge representation.
It is that a rule should not be trivial~(e.g. incomplete sentence or the latter sub-sentence simply repeats its former sub-sentence).


More concretely, we define the requirements for designing our framework as 1) there should be as fewer contradictions between facts and the rule as possible, and 2) the rule should reflect the reality, 3) the content in facts should be relevant specific statements that are covered by the rule, 4) the rule should not be trivial.

Based on this, we develop our framework as shown in Figure~\ref{fig:main_framework}. 
It consists of five modules, where module 1~(M1) is the rule proposer, module 2~(M2) is the deductive consistency evaluator, module 3~(M3) is the indiscriminate confirmation handler, module 4~(M4) is the generalization checker, and module 5~(M5) is the triviality detector.
Specifically, M1 is in charge of the generation of rules. 
M2, M3, M4, M5 are independent classification models each verifying rules with different requirement.
The role of M2/3/4/5 is similar to the verifier developed for deductive reasoning to make more solid reasoning steps~\citep{DBLP:journals/corr/abs-2205-12443}.
The independence of M2/3/4/5 makes it possible to run them in parallel.

In practice, we implement all five modules with PLMs. 
We call our implementation as CoLM~(\textbf{C}hain-of-\textbf{L}anguage-\textbf{M}odels).
The goal of M1 is to generate rules based on the input facts and a given rule template. 
Thus, M1's input contains facts, a rule template, and prompts that demonstrate the rule induction task.
M2 and M4's inputs include prompts that explain the rule-fact compatibility, a rule, and fact(s); M3 and M5's inputs include again prompts that explain the task and a rule, as their targets are independent of fact.

More interestingly, although our framework is solely based on the insights from philosophy literature, we also find a mathematical interpretation of this approach.
Here, we denote $P(A)$ as the probability indicating whether $A$ is valid for simplicity.
Thus, M2 and M4 jointly measure the validness of a fact given the corresponding rule
$P(fact|rule) \approx P_{M24}(fact|rule) = P_{M2}(fact|rule)P_{M4}(fact|rule)$, M3 and M5 directly measure the validness of the rule itself $P(rule) \approx P_{M35}(rule) = P_{M3}(rule)P_{M5}(rule)$.
Here $P_{M24}$ and $P_{M35}$ are parameterized as the product of two corresponding probabilities.
By using Bayes' rule, we can easily show that the validness of a rule based on the input fact is~(here we omit constant $P(facts)$)
\begin{equation}
\small
P(rule|fact)\approx{P_{M24}(fact|rule)P_{M35}(rule)}.
\end{equation}
Note that this score is merely a discrimination score and thus different from the generation probability from M1.
In other words, the rules proposed by M1 are then selected by M2/3/4/5 in a Bayesian inference fashion.

\section{Experiments}

In this section, we discuss the evaluation metrics and baselines, and then present the main results of our framework~(all are averaged by 5 runs).

\subsection{Evaluation Metrics}
We carry out evaluations for the framework~(the rule generation task with DEER) and individual modules for classification using DEERLET.

For evaluation of the rule generation of the overall framework, 
we use METEOR, WRecall, and GREEN as automatic evaluation metrics; And use precision, recall, f1, and the four metrics in DEERLET as human evaluation metrics.
WRecall, GREEN, and the four metrics in DEERLET are our newly proposed metrics for inductive reasoning introduced in \S\ref{adopted_proposed_automatic_evaluation_metrics}.

For evaluation of the classification tasks on DEERLET,
we use accuracy, f1, and averaged precision as metrics.
\begin{table*}[]
\centering
\resizebox{2\columnwidth}{!}{
\begin{tabular}{c|ccc|ccc|cccc}
\toprule
Models& METEOR & WRecall & GREEN& Precision~(\%) & Recall~(\%) & F1 & Consistent & Reality & General & Non-trivial \\ \midrule
R+F     & 11.20         & 0.50        & 2.37        & 9.0                                 & \textbf{100}                             & 0.17                  & 0.90                          & 0.15                           & 0.28                       & 0.85                       \\
M1      & 25.28         & 0.50        & 3.56        & 45.0                                & \textbf{100}                             & 0.62                  & 0.63                          & 0.60                           & 0.83                       & 0.86                       \\ \midrule
M1 + M2 & 25.68 / 25.69 & 0.53 / 0.54 & 3.68 / 3.71 & 45.9 / 59.8                         & 87.8 / 71.1                     & 0.60 / 0.65           & 0.63 / 0.75                   & 0.62 / 0.72                    & 0.83 / 0.92                & 0.86 / 0.94                \\
M1 + M3 & 25.39 / 26.57 & 0.50 / 0.59 & 3.57 / 3.95 & 45.2 / 60.2                         & 84.4 / 75.6                     & 0.59 / \textbf{0.67}           & 0.63 / 0.77                   & 0.60 / 0.74                    & 0.83 / 0.89                & 0.87 / 0.91                \\
M1 + M4 & 26.12 / 26.30 & 0.53 / 0.58 & 3.74 / 3.92 & \textbf{48.5} / 53.3                         & 92.2 / 88.9                     & \textbf{0.64} / \textbf{0.67}           & 0.64 / 0.67                   & \textbf{0.64} / 0.65                    & \textbf{0.84} / 0.91                & 0.88 / 0.89                \\
M1 + M5 & 25.28 / 25.76 & 0.50 / 0.54 & 3.55 / 3.74 & 46.1 / 48.1                         & 97.8 / 97.8                     & 0.63 / 0.65           & 0.64 / 0.66                   & 0.61 / 0.63                    & 0.83 / 0.83                & 0.88 / 0.91                \\
CoLM    & \textbf{26.44} / \textbf{27.32} & \textbf{0.54} / \textbf{0.62} & \textbf{3.78} / \textbf{4.11} & 48.1 / \textbf{70.0}                         & 72.2 / 54.4                     & 0.58 / 0.61           & \textbf{0.65} / \textbf{0.81}                   & \textbf{0.64} / \textbf{0.80}                    & \textbf{0.84} / \textbf{0.94}                & \textbf{0.90} / \textbf{0.97}               \\
\bottomrule
\end{tabular}}
\caption{Result of CoLM and baselines on DEER under in-context learning / finetuning setting.
The first three metrics are automatic metrics, and the last seven metrics are human evaluation metrics.
}
\label{tab:framework_result}
\end{table*}

\subsection{Baselines}
We use a non-neural method and a neural method as baselines for the framework.
We call the non-neural baseline ``R+F'', as it randomly fills the given rule template with sentences or phases from the given fact.
The neural baseline we use is the rule proposer itself in Figure~\ref{fig:main_framework}.

We use majority class and TF-IDF~\citep{DBLP:journals/jd/Jones04} as baselines for individual modules.
The majority class baseline always predicts ``yes'', which is equivalent to not using M2/3/4/5 to filter rules from M1.
TF-IDF is another reasonable baseline as the induced rules contain similar contents compared to input facts. 
In practice, each input fact-rule pair is assigned a TF-IDF value, and a threshold for correctness~(to compare with the TF-IDF value) is tuned on the DEERLET validation set. 

\subsection{Main Results}
Most modules are implemented with GPT-J~\citep{gpt-j}, a pre-trained language model with 6 billion parameters.
Results on other LLMs such as LLaMA~\citep{touvron2023llama}
can be found in \S\ref{appen:llama_vicuna}.
For better analysis, we conduct the experiments in two settings, including 
in-context learning setting~\citep{DBLP:journals/corr/abs-2107-13586,DBLP:conf/nips/BrownMRSKDNSSAA20}
and finetuning setting. 
The only exception is that we do not test finetuning setting on M1~(the only generative module), since we are mainly investigating~(out-of-box) PLM's ability.
However if with finetuning, language model might perform worse on out-of-distribution data and lose their generality for input facts from different topics~\citep{DBLP:conf/iclr/KumarRJ0L22}.
For this reason we do not implement with T5~\citep{DBLP:journals/jmlr/RaffelSRLNMZLL20}.

We report the results of in-context learning setting and finetuning setting in Table~\ref{tab:framework_result} and Table~\ref{exp:deerlet}.
The thresholds of M2/3/4/5 used in Table~\ref{tab:framework_result} and Table~\ref{exp:deerlet} are tuned on the DEERLET validation set.
More details on setting up thresholds are illustrated in \S\ref{appen:set_up_threholds}.
The results on DEER are shown in Table~\ref{tab:framework_result}.
As expected, the M1 alone outperforms the R+F baseline across the board, indicating that the PLM has some rule induction capability.
Augmenting the M1 with some filtering mechanism can reliably improve the generated rule quality further.
Lastly, our full model, CoLM, outperforms all baselines justifying the effectiveness of our proposed framework for natural language inductive reasoning.
%
Due to page limit, DEERLET results are analyzed in \S~\ref{appen:deerlet}.

\section{Analysis}

In this section, we investigate the question of ``how well can pretrained language models perform inductive reasoning?''.
Specifically, we provide analyses in terms of rule types, topics, variations of input fact, and scales of language models.
Except for Table~\ref{tab:input}, the input used is short fact, 3 fact, full fact.
Except for Table~\ref{fig:analysis_scale}, the model used is GPT-J.
All experiments in this section are based on the in-context learning setting, each averaged by 5 runs. Similar trends are also observed in other settings.
We report METEOR and GREEN as metrics in this section.
In addition to analyses with automatic evaluation results in this section, we also manually analyze the failure cases of CoLM in \S\ref{appen:failure_analysis}, by classifying error types and give a statistics on the percentage of the identified error types. 

\subsection{Different Rule Types}
Table~\ref{analysis:rule_templates} shows the breakdown evaluation of CoLM based on four basic rule types in formal language~\citep{DBLP:books/aw/RN2020}.
The mapping between the logic forms and corresponding natural language templates can be found in Table~\ref{tab:rule_template}.

\begin{table}[]
\resizebox{\columnwidth}{!}{
\begin{tabular}{c|cccc}
\toprule
Models & \thead{If \_\_,\\ then \_\_.} & \thead{There exists \_\_,\\ which \_\_.} & \thead{If \_\_ and \_\_,\\ then \_\_.} & \thead{If \_\_ or \_\_,\\ then \_\_.} \\ \midrule
R+F                        & 9.87 / 2.22       & 17.45 / 2.95                  & 10.63 / 2.30                & 12.53/ 2.50                \\
M1                         & 23.05 / 3.39      & 32.03 / 4.00                  & 27.01 / 3.67                & 29.09 / 3.81               \\ \midrule
M1+M2                      & 23.76 / \textbf{3.58}      & \textbf{33.13} / \textbf{4.39}                  & 26.00 / 3.43                & 28.76 / 3.69               \\
M1+M3                      & 23.34 / 3.46      & 31.35 / 3.80                  & 26.64 / 3.58                & 29.56 / 3.95               \\
M1+M4                      & 23.58 / 3.43      & 32.16 / 4.06                  & 25.94 / 3.48                & \textbf{29.80} / \textbf{4.05}               \\
M1+M5                      & 23.04 / 3.40      & 32.60 / 4.17                  & \textbf{27.05} / \textbf{3.68}                & 29.08 / 3.81               \\
CoLM                       & \textbf{24.15} / 3.55      & 32.50 / 4.16                  & 26.41 / 3.58                & 29.60 / 3.96               \\ \bottomrule
\end{tabular}}
\caption{Analysis of PLM~(GPT-J)'s performance (measured in METEOR / GREEN) in with different rule templates.}
\label{analysis:rule_templates}
\end{table}
The table shows that ``there exists \_, which \_'' achieves the best performance. 
It is reasonable, as simply copying the contents of facts to compose a rule will be acceptable for $\exists$ quantifier in logic.

\subsection{Different Topics}
Table~\ref{analysis:topics} shows the performance of CoLM over different topics. 
CoLM performs much worse on History and Physics than the other topics.
We attribute it to that the rules in history and physics have high variance, demand a higher level of abstraction, and are not very similar to the input facts. 
For example, in physics, many rules are natural language descriptions of physical laws such as Newton's law of universal gravitation, while the input facts might be the values of gravitational force and mass of specific objects.
In contrast, CoLM achieves better performance in Botany. 
One possible reason is that many rules in botany can be very similar to the input facts~(an example is shown in Table~\ref{tab:examples_inductive_reasoning}). 

\begin{table}[]
\centering
\resizebox{1\columnwidth}{!}{
\begin{tabular}{c|cccccc}
\toprule
Models & Zoology               & Botany                & Astronomy             & Geology               & History               & Physics               \\ \midrule
R+F                        & 9.65 / 2.20  & 10.24 / 2.26 & 13.09 / 2.56 & 13.28 / 2.58 & 11.07 / 2.35 & 11.44 / 2.39 \\
M1                         & 28.88 / 3.80 & 31.14 / 3.95 & 34.40 / 4.15 & 27.71 / 3.72 & 22.17 / 3.33 & 20.01 / 3.16 \\ \midrule
M1+M2                      & \textbf{29.70} / \textbf{4.00} & 30.59 / 3.76 & 32.88 / 3.82 & 28.67 / 4.08 & 22.65 / 3.50 & 20.49 / \textbf{3.30} \\
M1+M3                      & 29.17 / 3.85 & 31.03 / 3.88 & 33.86 / 4.04 & 28.16 / 3.87 & 22.30 / 3.36 & 20.16 / 3.17 \\
M1+M4                      & 29.00 / 3.77 & \textbf{31.54} / \textbf{4.06} & 34.17 / 4.20 & 28.63 / 4.04 & \textbf{25.00} / \textbf{3.89} & 20.16 / 3.22 \\
M1+M5                      & 28.72 / 3.76 & 31.26 / 3.99 & 34.60 / 4.21 & 27.33 / 3.62 & 22.01 / 3.26 & 20.00 / 3.10 \\
CoLM                       & 29.25 / 3.84 & 31.00 / 3.86 & \textbf{35.33} / \textbf{4.46} & \textbf{29.51} / \textbf{4.23} & 24.34 / 3.72 & \textbf{20.67} / \textbf{3.30} \\ \bottomrule
\end{tabular}}
\caption{Analysis of PLM~(GPT-J)'s performance (measured in METEOR / GREEN) in under different topics.}
\label{analysis:topics}
\end{table}

\subsection{Variations of Input Facts}

In table~\ref{tab:input}, long facts mean the paragraph-level facts in DEER, and short facts mean the core sentence-level facts selected from corresponding paragraph-level facts. 
The different number of facts indicates the different number of facts given as input that exhibit similar rule patterns ~(e.g. Lemon tree / orange tree / apple tree can conduct photosynthesis). 
We consider the number of facts as an important factor because psychological research shows that more facts with similar patterns can help with inductive reasoning~\citep{heit2000properties}.
Missing fact experiments are also conducted, where for each fact we randomly throw the former half or the latter half of the sentences. 
It is an important setting as it is hard for the input facts to cover all the elements of the desired rule in a realistic scenario.
As a result, it might be common that some required pieces of fact are missing.
The results indicate that larger number of concise but full facts are beneficial for rule induction, while too many facts with similar patterns might not be helpful.

\begin{table}[]
\centering
\resizebox{1.0\columnwidth}{!}{
\begin{tabular}{c|ccccc}
\toprule
Models & \thead{Long facts\\1 full facts} & \thead{Short facts\\ 1 full facts} & \thead{Short facts\\2 full facts} & \thead{Short facts\\3 full facts} & \thead{Short facts\\3 missing facts} \\ \midrule
R+F                        & 9.35 / 2.16         & 10.87 / 2.33          & 11.16 / 2.36          & 11.20 / 2.37          & 11.52 / 2.40                             \\
M1                         & 23.12 / 3.40        & 24.75 / 3.52          & 25.22 / 3.55          & 25.28 / 3.56          & 24.67 / 3.51                             \\ \midrule
M1+M2                      & 23.43 / 3.49        & 25.30 / 3.68          & 25.88 / 3.74          & 25.68 / 3.68          & 25.01 / 3.58                             \\
M1+M3                      & 23.25 / 3.44        & 24.91 / 3.55          & 25.32 / 3.57          & 25.39 / 3.57          & 24.77 / 3.52                             \\
M1+M4                      & 23.65 / 3.52        & 25.48 / 3.65          & 26.04 / 3.73          & 26.12 / 3.74          & 25.09 / 3.59                             \\
M1+M5                      & 23.23 / 3.44        & 24.81 / 3.54          & 25.31 / 3.58          & 25.28 / 3.55          & 24.81 / 3.57                             \\
CoLM                       & \textbf{24.03} / \textbf{3.60}        & \textbf{25.89} / \textbf{3.73}          & \textbf{26.71} / \textbf{3.85}          & \textbf{26.44} / \textbf{3.78}          & \textbf{25.41} / \textbf{3.65}                            \\ \bottomrule
\end{tabular}}
\caption{Analysis of PLM~(GPT-J)'s performance (measured in METEOR / GREEN) with different input lengths and whether fact contains enough information.}
\label{tab:input}
\end{table}

\subsection{Different Scales of PLMs}
\begin{figure}[t]
\centering
\resizebox{0.8\columnwidth}{!}{
\includegraphics[]{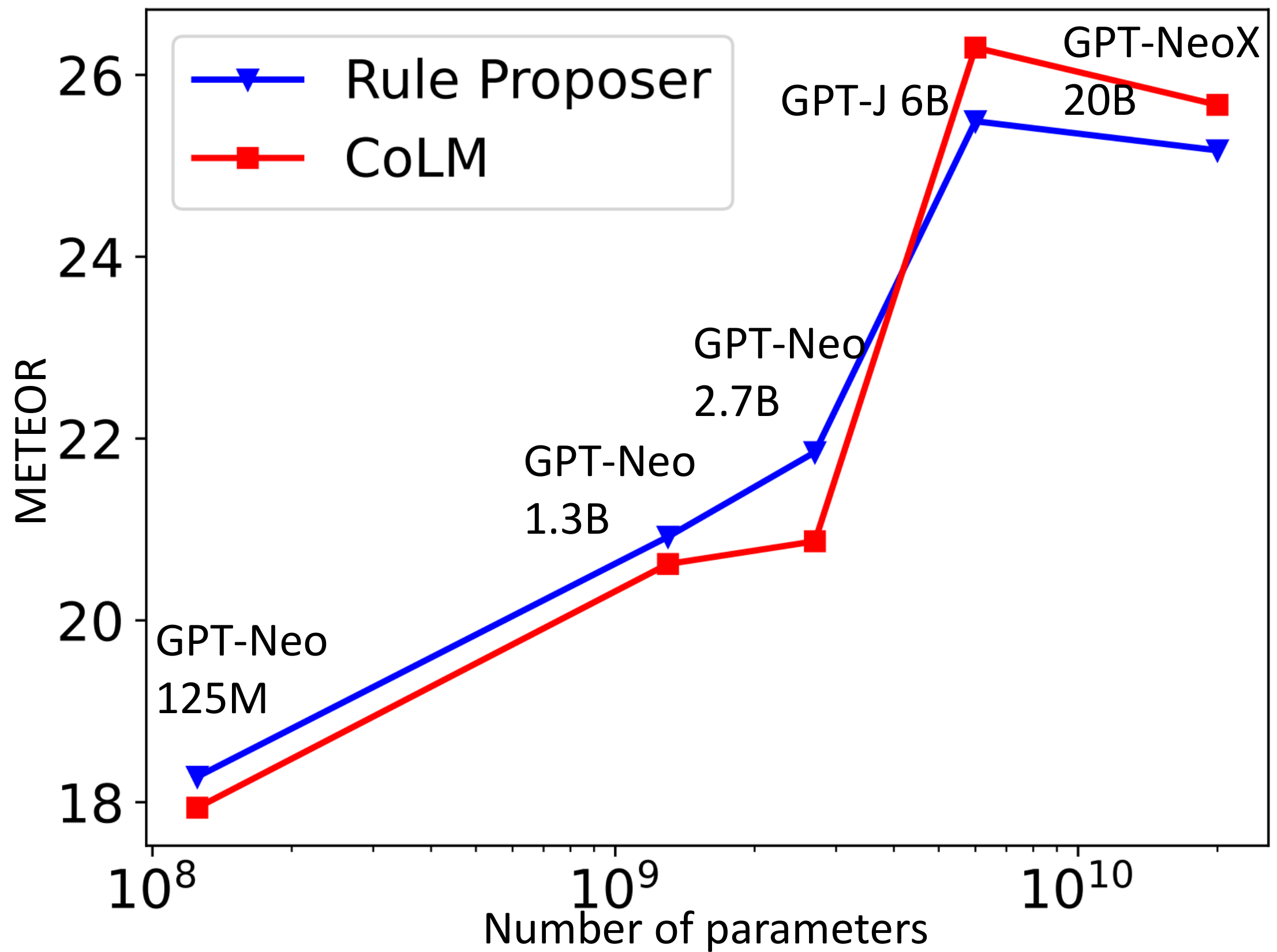}
}
\caption{Influence of the scale of PLM on inductive reasoning task with DEER~(measured with METEOR).
}
\label{fig:analysis_scale}
\end{figure}


Figure~\ref{fig:analysis_scale}
shows the influence of the scale of pre-trained language models~(under in-context learning setting) on induction.
Here, we consider GPT-Neo 125M, GPT-Neo 1.3B, GPT-Neo 2.7B, GPT-J 6B and GPT-NeoX 20B~\citep{gpt-j}.
The figure shows that generally performance of M1 steadily improves as the scale being larger, and M2/3/4/5 are only helpful since 6B parameters.
The only exception is that both M1 and M2/3/4/5 might reach a plateau in 20B parameters.

\subsection{Error Analysis}
\begin{figure}[t]
\centering
\resizebox{\columnwidth}{!}{
\includegraphics[]{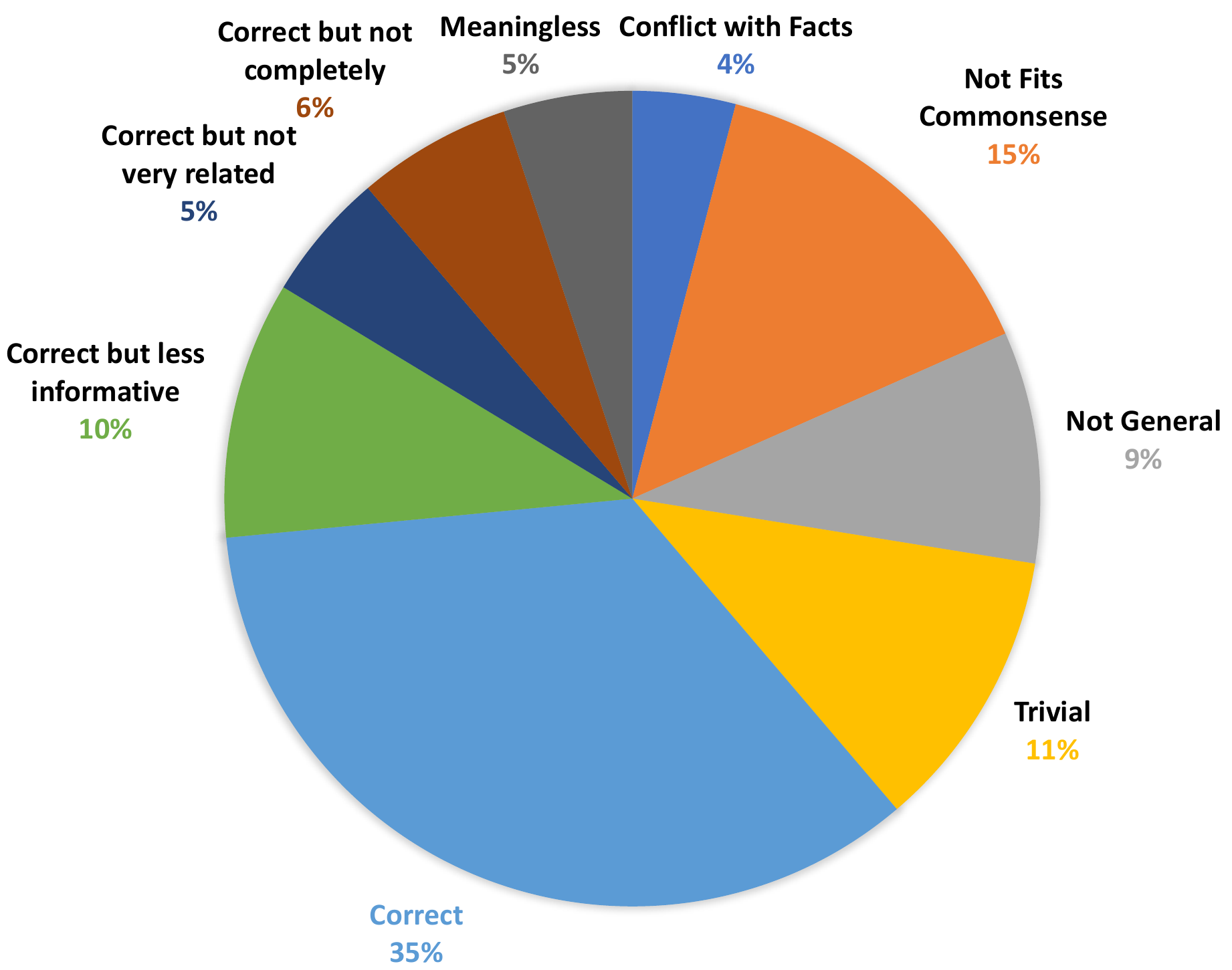}
}
\caption{Error Analysis of CoLM with finetuned Module 2/3/4/5. In total 100 rules are manually checked.
}
\label{fig:error_analysis}
\end{figure}

We sampled 100 rules from CoLM~(rules that generated by M1 and pass all M2/3/4/5), and have conducted an error analysis of the samples.
Figure~\ref{fig:error_analysis} shows the results.
Among them, ``Conflict with Facts'', ``Not Fits Commonsense~(not reflects reality)'', ``Not General'', and ``Trivial'' corresponds to the rules that should be filtered by CoLM but not.
We find that beyond ``Correct'' and errors made by classification modules, there are also some other classes that worth mentioning, but they could be seen as other kinds of ``Trivial''.
This figure shows that the four criteria we proposed are important for verification.
More details about error analysis can be found at \S~\ref{appen:failure_analysis}.

\FloatBarrier

\section{Overview and Future Perspectives of Inductive Reasoning}
\label{sec:overview_future_perspectives}

The first version of this paper was finished in 2022. At that time, inductive reasoning—in the sense of deriving explicit natural language hypotheses (rules) from observations (input facts), where the hypotheses and observations adhere to specific relations defined by induction—was a new and unexplored research area.

Previously, the most closely related works came from the ILP (Inductive Logic Programming) community, which focused on symbolic approaches to the task of inductive reasoning (inducing explicit formal language hypotheses). This paper aims to act as a bridge between the ILP and NLP communities by (1) demonstrating how natural language and related techniques (foundation models) can address challenges within the ILP community, and (2) introducing the definition and task of inductive reasoning to NLP. Moreover, this paper can serve as a preliminary study, suggesting that language models have the potential to function as inductive reasoners. The transcription of requirements for inductive arguments from philosophical literature, as illustrated in Section \ref{subsec:our_framework}, could remain useful even in the era of powerful LLMs.

The possible future challenges of research on inductive reasoning include (1) establishing and solving more challenging tasks for inductive reasoning, and (2) overcoming the fundamental challenges inherent in induction.

\subsection{Establishing and Solving More Challenging Tasks for Inductive Reasoning}

A naturally more challenging task is scientific hypotheses discovery, which is to generate novel and valid scientific hypotheses. Here, ``novel'' means ``not known or recognized by any literature''. In fact, inductive reasoning is one of the primary types of reasoning in the development of science. Essentially, scientists use inductive reasoning whenever they move from limited data to a more general conclusion~\citep{okasha2002philosophy}. Thus, exploring how to generate preliminary hypotheses~(a.k.a. research ideas) and possibly act as a ``copilot'' for scientists could be an intriguing research topic. \citet{DBLP:journals/corr/abs-2309-02726} extend inductive reasoning to the task of scientific hypothesis discovery, demonstrating that LLMs can generate novel and valid hypotheses in some social science disciplines. However, there are still many challenging questions to address, such as how to develope a system for other disciplines.

Another challenging task is pattern induction, which is to induce (executable) rules/patterns from complex (synthetic) facts. This task currently encompass (1) identifying patterns in a sequence of numbers~\citep{Qiu2023phenomenal}, (2) discerning arithmetic calculation patterns~\citep{zhu2023large}, and (3) detecting change patterns of 2D grid images~\citep{DBLP:journals/corr/abs-2309-05660}. 
The term ``executable'' is used here because many of these patterns can be described in the form of program. 
An advantage of pattern induction tasks is that challenging datasets can be efficiently constructed using synthetic methods.
This direction is also interesting as it can aid in understanding the inductive reasoning capabilities of LLMs and requires a combination of this understanding with the ability to generate program.

\subsection{Overcoming Fundamental Challenges Inherent in Induction}
This challenge stems from certain fundamental requirements for the induced rules. As illustrated in Section \ref{subsec:our_framework}, some of these requirements include
\begin{itemize}
    \item Checking whether the induced rule accurately reflects reality.
    \item Determining whether the hypotheses are more general than the observations.
\end{itemize}

Here, the ``reflects reality'' in the first requirement refers to whether the rule mirrors the objective world (or the environment of the task). In certain task settings, such as scientific hypothesis discovery, verifying whether an induced hypothesis mirrors the objective world can be very challenging, given that LLMs do not directly interact with the world. To ascertain the validity of the hypotheses, LLMs might need to utilize tools to conduct actual experiments to test the induced hypotheses. In other tasks, such as pattern induction, meeting this requirement could be much simpler, as whether it catches the designed patterns can be examined by executing the program and checking whether it produces the expected output.

The second requirement can be interpreted as ``whether the hypothesis is novel compared to the all existing literature'' in the task of scientific hypothesis discovery~\citep{DBLP:journals/corr/abs-2309-02726}. Meeting this requirement involves key challenges including information retrieval and novelty checking. 

\section{Conclusion}

To overcome the systematic problems of using formal language for inductive reasoning, we propose a new paradigm~(task) of inducing natural language rules from natural language facts, and correspondingly propose a dataset DEER and new evaluation metrics for this task.
We provide the first and comprehensive analysis of PLM's ability to induce natural language rules from natural language facts.
We also propose a new framework, drawing insights from philosophical literature, which, as shown in the experimental section, surpasses baselines in both automatic and human evaluations.

\section*{Limitations}
In this work, the size of dataset~(DEER) contains 1.2k fact-rule pairs, which is relatively small.
The reason is that the ``rules'' in this task are required to be very general. It is not easy to collect a large set of such rules in high-quality.
Additionally, a rule can be collected only if (1) there are several facts findable in online texts, and (2) these facts satisfy certain relation with the rule required by induction (the rule generalizes over the facts).

In addition, the DEER dataset mainly covers commonsense knowledge.
A successive work to this paper~\citep{DBLP:journals/corr/abs-2309-02726} focuses on a more challenging setting of inductive reasoning, which is to generate novel and valid scientific hypotheses~(e.g., Newton's Laws are scientific hypotheses). 
Here novel is defined as ``not known or recognized by any literature'', which means this new setting is very challenging even for the most advanced LLMs.

\section*{Acknowledgments}
This research/project is supported by the Ministry of Education, Singapore under its MOE Academic Research Fund Tier 2 (STEM RIE2025 Award MOE-T2EP20123-0005).

We sincerely appreciate the anonymous reviewers who have given careful reviews to this paper, and the anonymous chair who have closely looked into the this paper. 

\bibliography{custom}

\FloatBarrier
\appendix

\section{Appendix}
\label{sec:appendix}

\subsection{Hyperparameters}
For finetuning experiments, we use learning rate 1e-5; weight decay 0.1; adam epsilon 1e-8; batch size 4; and early stopping with accuracy as the metric. 
We perform our experiments on RTXA6K GPU.
We use nltk package to calculate BLEU and METEOR.

\subsection{DEERLET Results}
\label{appen:deerlet}

\begin{table}[]
\centering
\resizebox{1.0\columnwidth}{!}{
\begin{tabular}{c|ccc}
\toprule
Metrics        & Accuracy (\%)      & F1              & Averaged Precision    \\ \toprule
               & \multicolumn{3}{c}{\textbf{Deductive Consistency Evaluator (M2)}}     \\ \midrule
Majority class & 62.5               & 0.77           & 0.63                  \\
TF-IDF         & 62.5               & 0.77           & 0.69                  \\ \midrule
GPT-J          & 61.5 / \textbf{74.0}        & 0.71 / \textbf{0.83}     & 0.75 / \textbf{0.83}           \\ \toprule
               & \multicolumn{3}{c}{\textbf{Indiscriminate Conformation Handler (M3)}} \\ \midrule
Majority class & 60.0               & 0.75           & 0.60                  \\
TF-IDF         & 60.0               & 0.75           & 0.64                  \\ \midrule
GPT-J          & 56.0 / \textbf{70.5}        & 0.57 / \textbf{0.77}     & 0.66 / \textbf{0.79}           \\ \toprule
               & \multicolumn{3}{c}{\textbf{Generalization Checker (M4)}}              \\ \midrule
Majority class & 83.0               & 0.91            & 0.83                  \\
TF-IDF         & 83.0               & 0.91            & 0.86                  \\ \midrule
GPT-J          & 71.0 / \textbf{86.0}        & 0.82 / \textbf{0.92}     & 0.87 / \textbf{0.97}           \\ \toprule
               & \multicolumn{3}{c}{\textbf{Triviality Detector (M5)}}                 \\ \midrule
Majority class & 86.0               & 0.93            & 0.86                  \\
TF-IDF         & 86.0               & 0.93            & 0.90                  \\ \midrule
GPT-J          & 78.5 / \textbf{89.5}        & 0.87 / \textbf{0.94}     & 0.89 / \textbf{0.94}           \\ 
\bottomrule
\end{tabular}}
\caption{Results on DEERLET for different modules under in-context learning / finetuning settings.}
\label{exp:deerlet}
\end{table}

The results on DEERLET are summarized in Table~\ref{exp:deerlet}. 
In this experiment, we investigate the classification performance of language models in terms of different aspects required by inductive reasoning, which includes deductive consistency, indiscriminate confirmation, and generalization / triviality classification. 
%
It shows that TF-IDF achieves the same performance with majority class baseline in accuracy and f1 metrics.
The reason is that the best thresholds obtained for TF-IDF are all zero, which means that TF-IDF value is not effective for the four tasks.
It also shows that with in-context learning GPTJ performs worse than the majority class baseline, while finetuned GPTJ steadily performs better.

\subsection{Failure Analysis}
\label{appen:failure_analysis}

We sampled 100 rules from CoLM~(rules that generated by M1 and pass all M2/3/4/5), and have conducted an error analysis of the samples.
Figure~\ref{fig:error_analysis} shows the results.

Among them, ``Conflict with Facts'', ``Not Fits Commonsense~(not reflects reality)'', ``Not General'', and ``Trivial'' corresponds to the rules that should be filtered by CoLM but not.
However, we find that beyond ``Correct'' and errors made by classification modules, there are also some other classes that worth mentioning.

``Correct but less informative'' means some facts that is not trivial~(by our former description of triviality -- incomplete sentences or the conclusion simply repeats some part of premises.), not incorrect, but not very informative.
Examples include ``if a bird can help a plant to reproduce, then it is probably a good thing for the plant'', and ``if a land is green, then it probably contains forests''.

``Correct but not very related'' means although the rule is correct, but it is not very related to the facts given.
For example, the facts are only about the depth and shape of Marianas Trench, while the rule is ``if there exists a place with a greater depth, then it is possible to find something strange and interesting''~(the ``find something strange and interesting'' aspect is not mentioned in facts).

``Correct but not completely'' means the rule is somewhat to mostly correct, such as ``if a fruit has a strong smell, then it probably tastes good''~(while facts are about durian, champedek, and morinda citrifolia);  ``if an economy is based on textiles, then it might experience an industrial revolution''~(this rule is only true during a specific period of time in history); ``if a wire moves, then it might induce voltage in the conductor''~(this rule is only true if given magnetic fields).

``Meaningless'' means the rule is from a strange angle and it's hard to justify whether it is correct or not, such as ``if an event has a positive impact on an individual and on family, then the impact on the family is greater'', and ``if a man has experienced hardships and life has been tough, then he might be able to understand and change his ways in the future''.

\subsection{More Details on Difference with Other Reasoning Tasks}

In this paper, we strictly follows the definition and categorization of logical reasoning~(including deductive, inductive, and abductive reasoning) in a survey of logical reasoning~\citep{DBLP:journals/corr/abs-2303-12023}.

There have been some NLP works on case-based reasoning~\citep{DBLP:conf/emnlp/DasZTGPLTPM21,DBLP:conf/icml/DasGNTZHJM22,DBLP:conf/eacl/YangDCC23}, which can also be seen as inductive reasoning.
However, CBR is a different inductive reasoning type than the ``generalization'' process~(from facts to rules) described in~\citet{flach2000abductive}, but more on the general description on inductive reasoning~\citep{salmon1989introduction} that premises cannot conclusively provide support to the conclusion.

Inductive reasoning is also different from commonsense reasoning~\citep{DBLP:conf/emnlp/YangDRC20}, where commonsense reasoning focuses more on the ``knowledge'' aspect, and inductive reasoning focuses more on the ``reasoning'' aspect~\citep{DBLP:journals/corr/abs-2303-12023}.

\subsection{Annotation Details for DEERLET}
\label{appen:annotation_detail_deerlet}

In DEERLET, given fact(s) and a rule, the annotation targets are whether the rule satisfies four requirements.

Specifically, the requirements are ``if the rule is deductively consistent with the fact'', ``if the rule reflects reality'', ``if the rule is more general than the fact'', and ``if the rule is not trivial''.

The first three requirements are annotated on a 3-point scale~(true / partially true / false), and the last is annotated on a 2-point scale~(true / false).

Here we explain the standards of annotation on the four requirements. 

For ``if the rule is deductively consistent with the fact'', 
a 2-point will be assigned if the rule is totally relevant and consistent with the facts;
a 1-point will be assigned if the rule introduces new information that does not show in facts but is consistent with the given fact as well as some limited amount of commonsense knowledge related to the facts;
a 0-point will be assigned if the rule is (1) in conflict with given facts or (2) totally irrelevant to given facts or (3) introduces new information that is obviously wrong.

For ``if the rule reflects reality'', 
a 2-point will be assigned if the rule totally reflects reality;
a 1-point will be assigned if the rule reflects reality at most of the time;
a 0-point will be assigned if (1) the rule is totally incorrect or (2) the rule is only occasionally correct.

For ``if the rule is more general than the fact'',
a 2-point will be assigned if (1) the rule is more general than the facts or (2) it is obvious that the rule is trying to be more general than the facts;
a 1-point will be assigned if (1) it is even hard for humans to induce a more general rule from the given facts or (2) the rule copies part of the given facts that are already containing very general information;
a 0-point will be assigned if (1) from the facts it's easy for humans to induce a more general rule but the rule is not more general or (2) the rule is totally irrelevant to the facts.

For ``if the rule is not trivial'',
a 0-point will be assigned if (1) the rule is an incomplete sentence or (2) the latter sub-sentence of the rule only repeats the information in the former sub-sentence of the rule;
otherwise, a 1-point will be assigned.

\subsection{METEOR or GREEN?}
\label{appen:meteor_or_green}

Since inductive reasoning over natural language is a new task, and new metrics are designed~(e.g., WRecall, GREEN), it is important to understand which aspects each metric focus on and which metric should we pay more attention to.

As mentioned in \S\ref{adopted_proposed_automatic_evaluation_metrics}, METEOR can be seen as evaluating the ``precision'' of the final rules, while GREEN evaluates ``precision'' and ``recall'' at the same time.

However, it should be aware that the ``recall'' here is not as important as the ``recall'' in other tasks.
More specifically, here ``recall'' measures how many good rules generated by M1 are filtered by M2/3/4/5. 
However, we can use M1 to generate a large number of rules, and as long as CoLM has good precision, it is easy to obtain a large number of high-quality rules, especially considering that the computational cost of only inference of M1 is relatively very low. 

Based on this observation, we argue that ``precision'' should be a much more important aspect of evaluation compared to ``recall''~(measured by WRecall) or even ``f1''~(measured by GREEN) for this task.
More specifically, ``recall'' can be used to mainly measure at what efficiency can the system obtain rules with high precision.

This viewpoint of evaluation metrics, of course, can raise the question of whether some typical kinds of rules are mostly filtered when pursuing rules with high precision, and in the end inductive reasoning system with high precision might only be able to obtain some other typical kinds of rules. 
We leave this question as an open question for this task to solve in the future.

\subsection{Why METEOR not BLEU}
\label{appen:why_meteor_not_bleu}

We choose METEOR since METEOR has a higher correlation coefficient with human evaluation than BLEU.

More specifically, on DEERLET, we calculate the METEOR and BLEU for each generated rule with its golden rule in DEER and collect the human evaluation for the generated rule from label0/1/2/3 annotations in DEERLET~(we normalize each label to [0,1] and use the product of label0/1/2/3 as the overall human evaluation score for the generated rule).
Then, we can calculate the correlation coefficient between METEOR / BLEU and the overall human evaluation score.

On DEERLET, the correlation coefficient between METEOR and human evaluation is 0.29, it is statistically significant as its p-value is $4.48*10^{-6}$, smaller than the significance level~(0.05).
Similarly, the correlation coefficient between BLEU and human evaluation is 0.24, with p-value of $1.17*10^{-72}$, which is also significant.

We called 0.29 relatively high since in other open-ended NLP tasks such as dialogue systems, the Pearson correlation is typically only around 0.14~0.19 (shown in Table 3 in \citep{DBLP:conf/emnlp/LiuLSNCP16}, BLEU’s Pearson correlation is lower than METEOR’s in most of the time). However recent papers published in ACL 2023 on dialogue systems still adopt METEOR or BLEU as automatic evaluation metrics~\citep{DBLP:conf/acl/0002Z23,DBLP:conf/acl/0007YLRVC23,DBLP:conf/acl/LiLGL23}.


Developing better metrics for measuring the similarity between sentences is a challenging topic in NLP. 
Of course, METEOR is not a ``perfect'' automatic evaluation metric for inductive reasoning. 
We leave the question of ``what is a better metric for inductive reasoning over natural language'' as an open question for future works in the field.

One good thing is that WRecall and GREEN can be applied with many metrics measuring sentence similarity such as METEOR and BLEU, so the evaluation of ``recall'' should be able to also benefit from the advance of metrics that evaluate ``precision''.

\subsection{Difficulty in Designing Automatic Evaluation Metrics for Inductive Reasoning Tasks and How Should We Interpret the Results of Automatic Metrics}
\label{appen:difficulty_automatic_metrics}
Designing automatic evaluation methods for inductive reasoning is fundamentally difficult, mainly because of two reasons. 
Firstly, generalizing over existing facts is not restricted in a single way. Given existing facts, multiple rules that are very diverse from each other could all be true.
Secondly, when it comes to more difficult inductive reasoning data, it is nearly inevitable to use long sentences for facts and rule, which make it even harder for common evaluation metrics such as BLEU or METEOR.

However, we argue that although we don't have perfect automatic evaluation metrics for inductive reasoning now, it is not a reason to stop exploring research on inductive reasoning.
In fact, with the fast development of LLMs, more difficult tasks are needed to further explore the scientific boundary in NLP, and many recently proposed tasks are so difficult to be evaluated with automatic evaluation metrics that they fully rely on human evaluation~\citep{zhong2023goal,wang2023learning}.
In terms of human evaluation metrics, we also have proposed meaningful human evaluation metrics for inductive reasoning tasks shown in the last four columns in Table~\ref{tab:framework_result}, which are derived from philosophy literature~(the four requirements for induced rules, and the four requirements are also used to develop the CoLM framework).

The reason we try to propose suitable automatic evaluation metrics is that we hope to simplify the evaluation process for the inductive reasoning task~(at least for preliminary evaluations).
We have illustrated why these metrics should be reasonable in \S\ref{appen:meteor_or_green} and \S\ref{appen:why_meteor_not_bleu}.
Similar to inductive reasoning, abductive reasoning also have multiple diverse correct generations, however abductive reasoning generation task also utilizes METEOR or BLEU~\citep{DBLP:conf/iclr/BhagavatulaBMSH20} as automatic metrics.
In the future, the automatic metrics are possible to be further improved with the help of the community.
While for now, just like other recent difficult tasks~\citep{zhong2023goal,wang2023learning}, human evaluations are always preferred, but automatic evaluation metrics, though not perfect, can still be used as a fast evaluation metrics that can provide some insights for experiments.

\subsection{Results on Other LLMs}
\label{appen:llama_vicuna}

Table~\ref{tab:llama} shows the results of CoLM using LLaMA, under in-context learning setting.
Overall, CoLM outperforms all baselines, but the gap between M1 and CoLM are smaller.
The reason is that LLaMA tends to generate very sound rules, thus the M2/3/4/5 of CoLM barely filter any rules. 
Therefore the results of CoLM and M1 are closer. 
We think there are two reasons:
(1) with the fast development of LLMs, our proposed dataset is less challenging for more recent LLMs such as LLaMA;
(2) M2/3/4/5 instantiating with LLaMA have not been finetuned, but just in-context learning setting. 
Given that finetuned GPT-J largely improves GPT-J under in-context learning setting in Table~\ref{tab:framework_result}, a finetuned LLaMA should be able to filer more unreasonable generations.

While our work takes the first step to inductive reasoning in NLP and provide the first analysis, introducing more challenging inductive reasoning benchmarks would be beneficial to the the further development of the inductive reasoning field in NLP.


\subsection{Meaning of ``More General'' Required by Inductive Reasoning}
\label{appen:general_inductive_reasoning}





Given an argument consisting of a premise and a conclusion, if the conclusion involves new information that is not covered by the premise and can not be conclusively entailed by the premise, the argument is an inductive argument~\citep{salmon1989introduction}.

When the conclusion has a larger scope of information coverage than the premise, and can entail the premise, it can be said that the conclusion is ``more general'' to the premise. 
In this case, we termed the premise as a ``fact'', and the conclusion as a ``rule'';
When the conclusion contains new pieces of information and cannot entail the premise, as defined by~\citet{salmon1989introduction}, the argument is still an inductive argument.
But in this case, we termed the premise as a ``fact'', and the conclusion as another ``fact''.

For instance, if facts that are about cats and dogs are good accompaniment of humans, then some examples of a ``more general'' rule can be (1) mammals are good accompaniment of humans, or (2) domesticated animals are good accompaniment of humans, or (3) animals with four legs are good accompaniment of human.

In these examples, the rules cover a larger scope than the facts~(e.g., mammals compared to cats; domesticated animals compared to cats), and therefore the rules are ``more general'' than the facts.

``More general'' means not only about finding higher taxonomic rank, but can be in unlimited forms.
For instance, if the fact is about the Sun rises and falls every day, then some examples of a ``more general'' rule can be (1) the Earth is the king of the universe or (2) the Earth is rotating itself.

Both rule examples are ``more general'' than the given fact, since the rule can entail not only the given fact, but also other not mentioned facts such as the observable movements of the other stars in the Milky Way.


\begin{table}[]
\centering
\resizebox{0.4\columnwidth}{!}{
\begin{tabular}{ll}
\toprule
Model                      & LLaMA-7B     \\ \midrule
R+F                        & 11.20 / 2.37 \\
M1                         & 24.94 / 3.53 \\ \midrule
M1+M2                      & 25.12 / 3.54 \\
M1+M3                      & 24.77 / 3.49 \\
M1+M4                      & 25.42 / 3.60 \\
M1+M5                      & 25.74 / 3.68 \\
CoLM                       & {\bf 29.37} / {\bf 3.95} \\ 
\bottomrule
\end{tabular}}
\caption{In context learning results of LLaMA, measured in METEOR and GREEN.}
\label{tab:llama}
\end{table}
\subsection{Set up Thresholds for M2/3/4/5}
\label{appen:set_up_threholds}

Setting up thresholds is an important step for our framework, since different thresholds can lead to different inductive reasoning results. 
We discuss the details of setting up thresholds in the section.

We design the standard for setting up thresholds based on heuristics that the thresholds should be set up that each module~(in M2/3/4/5) should filter some rules but a single module should not filter too many rules~(in this case, since we have many modules, there might not remain a reasonable proportion of rules left). 

More specifically, given a rule (and facts), M2/3/4/5 can produce a score on evaluating the validity of the rule from a specific aspect.
The score is the ratio of the probability of the ``yes'' token and ``no'' token obtained from the last layer of PLM.
The score is in the range of [0,1].

We find that getting a specific threshold for each module is more beneficial than using the default 0.5 threshold.
We obtain the thresholds on the DEERLET validation set.

More concretely, on the validation set,
if there exists a global optimal threshold that (1) achieves the best f1 or accuracy and (2) the threshold should not be very close to 0 or 1 and (3) recall is not very close to 0~(when close to 1, it should not be in the case that the threshold accepts nearly all generated rules but should be that the threshold already rejects some rules), then the global optimal threshold is adopted;
if there is no such global optimal threshold, then find a local optimal threshold that (1) achieves the best f1 or accuracy compared to its neighboring thresholds and (2) the threshold should not be very close to 0 or 1, and (3) the recall range is in [0.7, 0.9], then the local optimal threshold is adopted.

\subsection{More Details to Prevent Collection of Generated Trivial Rules}
We use a simple heuristic method to prevent collection of generated trivial rules.
Specifically, only rules generated from Module 1 that is with more than 45 tokens~(not 45 words) do we pass to it Module 2/3/4/5, otherwise we directly filter it.

The reason that we set it up is that we find generated rules with less than 45 tokens are mostly~(if not all) incomplete sentences.
If we collect and label these incomplete sentences to finetune Module 2/3/4/5, then Module 2/3/4/5 mostly learn to classify whether the rules are complete or not, but not to learn the designed patterns~(since the label0/1/2/3 in DEERLET for incomplete sentences are all false).

For this reason, all annotated data in DEERLET only use rules that contain at least 45 tokens.

\subsection{Related Works on Inductive Logic Programming}
\label{appen:inductive_logic_programming}
Inductive Logic Programming~(ILP) is a subfield of machine learning that uses FOL to represent hypotheses and data. 
It relies on formal language for knowledge representation and reasoning purposes~\citep{de2010inductive}.
We propose a new paradigm that can naturally avoid three systematic disadvantages of ILP~\citep{DBLP:journals/ml/CropperDEM22}.
\citet{DBLP:journals/ml/CropperDEM22} summarizes the challenges for ILP, including disability of handling raw input such as natural language and image, sensitiveness to mislabeled data and incapacity to handle ambiguous input.
In this work, we propose a new paradigm/ for inductive reasoning to use natural language as representation for knowledge and PLM as inductive reasoners, which can naturally avoid these challenges.

Recently, \citet{DBLP:conf/ijcai/DaiM21} propose to use logic programming to induce knowledge form image raw input. 
Our work instead focus on natural language raw input, and use PLMs as reasoning methods to induce knowledge.

\subsection{Induce Rules from General Facts and Specific Facts}

Sixty percent of the rules in DEER are more general than any of their facts alone at least in one dimension. 
We describe this process as ``inducing general rules from specific facts''.
However, we find that there are many general statements (also referred to as general fact) of a rule on the web.
Therefore, for rule induction systems to be able to utilize both ``specific facts'' and ``general facts'', forty percent of the rules in DEER are equipped with general facts.
We describe this process as ``inducing general rules from general facts''.


Table~\ref{analysis_specific_general} shows the result from specific vs general facts under in-context learning and finetuning settings correspondingly. 
We have discussed that a rule induction system would be more widely applicable if it can utilize both specific fact and general fact.
In table~\ref{analysis_specific_general}, general facts cases result in lower performance. 
We think one of the most possible reasons is that in DEER many general facts do not directly contain the content of the corresponding gold rules. 
For example, general facts can be mottos from philosophers such as Socrates, and rules can be an understandable description of such mottos in natural language rule format.


\begin{table}[]
\centering
\resizebox{0.65\columnwidth}{!}{
\begin{tabular}{c|cc}
\toprule
Models & Specific facts & General facts \\ \midrule
R+F                        & 10.15 / 2.25   & 12.79 / 2.53  \\
M1                         & 25.61 / 3.58   & 24.57 / \textbf{3.51}  \\ \midrule
M1+M2                      & 26.47 / 3.82   & 24.14 / 3.42  \\
M1+M3                      & 25.88 / 3.64   & 24.38 / 3.45  \\
M1+M4                      & 27.19 / 3.91   & 24.36 / 3.48  \\
M1+M5                      & 25.59 / 3.57   & \textbf{24.61} / \textbf{3.51}  \\
CoLM                       & \textbf{27.74} / \textbf{3.98}   & 24.34 / 3.47  \\ \bottomrule
\end{tabular}}
\caption{Analysis of PLM~(GPT-J)'s performance (measured in METEOR / GREEN) in with specific or general input facts~(Under in-context learning setting).}
\label{analysis_specific_general}
\end{table}




\subsection{GPT3's Performance as Rule Proposer}
\begin{table}[]
\centering
\resizebox{1\columnwidth}{!}{
\begin{tabular}{c|ccccc}
\toprule
Models & Ada  & Babbage & Curie & GPTJ & Davinci \\ \midrule
R+F    & 1.21 & 1.81    & 1.88  & 1.86 & 1.86    \\ 
M1     & 5.41 & 4.29    & 5.76  & 4.00 & 7.52    \\ \bottomrule
\end{tabular}}
\caption{GPT-3's performance as well as GPT-J's performance as Rule Proposer~(Measured in BLEU).}
\label{appen:gpt3}
\end{table}
\label{appen:section:gpt3}

Table~\ref{appen:gpt3} shows the result to use GPT-3 and GPT-J as rule proposer~(M1). 
It is measured in BLEU because it's a very early result, and we haven't adopted METEOR yet.
If use METEOR as metric, the trend should be similar~(the trend of BLEU and METEOR are very similar in our other experiments).
The reason we do not test the scale performance of CoLM compared to M1 is that OpenAI's API does not support return full embeddings, and our current code relies on embedding to implement M2/3/4/5 of CoLM.
We will modify our code and try it on GPT-3 in the next version of our paper.

\subsection{Method for Prevention of Personal Information}
The first author collected the datasets. 
During collection,
(1) most of the data are collected from Wikipedia, where personal information is nearly none;
(2) the first author checks the data first before collects them.

\subsection{Prompt for ALL Modules}
We have uploaded the full code to GitHub, containing the full prompts. 
The full prompts can be also found in the uploaded supplementary materials along with this submission in utils.py.


\subsection{Dataset Split of DEER and DEERLET}
Out of the 1,200 rule-fact paris of DEER, 438 / 762 are designed for train / test. 
Out of 846 examples of DEERLET, 546 / 100 / 200 are designed for train / val / test.

In our previous arXiv version, we use a different dataset split (train 100 rules / test 100 rules), the current dataset split is (train 73 rules / test 127 rules) to better utilize the data (each rule has 6 annotated facts). The last 22 rules in test set (id: 105~126) are inspired by gpt-3.5-turbo, while all other rules are proposed by an expert. All facts are existing texts collected from the web using search engine, after given a rule.

\subsection{More Illustration on Human Evaluation}
Here the human annotations for human evaluation in Table~\ref{tab:framework_result} are from the DEERLET annotations.
DEERLET is annotated by an expert~(the first author).
The dataset~(DEERLET) is annotated before M2/3/4/5~(full CoLM) or any baseline experiments, so that the human evaluation is not influenced by the performance of any specific method.

More details about the DEERLET annotation are illustrated in \S\ref{appen:annotation_detail_deerlet}.

\end{document}